%% file: main.tex
\documentclass[conference]{IEEEtran}
\usepackage{amsmath,amssymb,amsfonts}
\usepackage{algorithmic}
\usepackage{algorithm}
\usepackage{graphicx}
\usepackage{fixltx2e}
\usepackage{stfloats}
\usepackage{placeins}
\usepackage{multirow}
\usepackage{booktabs}

\usepackage{acronym}
\input{./chapter/acronyms.tex}

\input{./chapter/commands.tex}

\usepackage[
  letterpaper,
  left=0.67in,
  right=0.67in,
  top=0.75in,
  bottom=1.0in,
  columnsep=0.25in
]{geometry}
	
\usepackage[%
	backend=biber,%
	style=ieee,%
	isbn=false,%
	hyperref=true,%
	maxbibnames=99,%
	sorting=none,%
	natbib=true,%
	language=english,%
	]{biblatex}%
\DeclareFieldFormat{sentencecase}{\csname bbx@colon@search\endcsname#1}

\def\BibTeX{{\rm B\kern-.05em{\sc i\kern-.025em b}\kern-.08em
    T\kern-.1667em\lower.7ex\hbox{E}\kern-.125emX}}

\usepackage{tikz}
\usepackage{textcomp}
\usepackage{hyperref}
\usepackage{lipsum}

\newcommand\copyrighttext{%
  \footnotesize \textcopyright \the\year{} IEEE. Personal use of this material is permitted. Permission from IEEE must be obtained for all other uses, including reprinting/republishing this material for advertising or promotional purposes, collecting new collected works for resale or redistribution to servers or lists, or reuse of any copyrighted component of this work in other works.}
\newcommand\copyrightnotice{%
\begin{tikzpicture}[remember picture,overlay]
\node[anchor=south,yshift=10pt] at (current page.south) {\fbox{\parbox{\dimexpr\textwidth-\fboxsep-\fboxrule\relax}{\copyrighttext}}};
\end{tikzpicture}%
}

\begin{document}

\title{Robust Semi-Supervised Temporal Intrusion Detection for Adversarial Cloud Networks

}



\author{\IEEEauthorblockN{Anasuya Chattopadhyay}
\IEEEauthorblockA{German Research Center for\\Artificial Intelligence (DFKI),\\
Germany\\
Email: anasuya.chattopadhyay@dfki.de}
\and
\IEEEauthorblockN{Daniel Reti}
\IEEEauthorblockA{German Research Center for\\Artificial Intelligence (DFKI),\\
Germany\\
Email: daniel.reti@dfki.de}
\and
\IEEEauthorblockN{Hans D. Schotten}
\IEEEauthorblockA{German Research Center for\\Artificial Intelligence (DFKI),
and\\
RPTU University Kaiserslautern-Landau, Germany\\
E-mail: schotten@eit.uni-kl.de}}

\maketitle
\copyrightnotice

%
%
%
%
\begin{abstract}
Cloud networks increasingly rely on machine learning based Network Intrusion Detection Systems to defend against evolving cyber threats. However, real-world deployments are challenged by limited labeled data, non-stationary traffic, and adaptive adversaries. While semi-supervised learning can alleviate label scarcity, most existing approaches implicitly assume benign and stationary unlabeled traffic, leading to degraded performance in adversarial cloud environments. This paper proposes a robust semi-supervised temporal learning framework for cloud intrusion detection that explicitly addresses adversarial contamination and temporal drift in unlabeled network traffic. Operating on flow-level data, this framework combines supervised learning with consistency regularization, confidence-aware pseudo-labeling, and selective temporal invariance to conservatively exploit unlabeled traffic while suppressing unreliable samples. By leveraging the temporal structure of network flows, the proposed method improves robustness and generalization across heterogeneous cloud environments. Extensive evaluations on publicly available datasets (CIC-IDS2017, CSE-CIC-IDS2018, and UNSW-NB15) under limited-label conditions demonstrate that the proposed framework consistently outperforms state-of-the-art supervised and semi-supervised network intrusion detection systems in detection performance, label efficiency, and resilience to adversarial and non-stationary traffic.

\end{abstract}

\begin{IEEEkeywords}
Network intrusion detection systems, Cloud computing security, Semi-supervised learning, Adversarial machine learning, Network security, Deep learning
\end{IEEEkeywords}

%
%
%
%

 \section{Introduction} 
 \input{./chapter/introduction.tex}


 \section{Related Work} 
 \input{./chapter/works.tex}


\section{Methodology} 
\input{./chapter/method.tex}


\section{Experimental Setup} 
\input{./chapter/setup.tex}


\section{Results and Discussion} 
\input{./chapter/results.tex}

%
%
%
%
\section{Conclusion and Outlook}
\input{./chapter/conclusion.tex}

%
%
%
%
\section*{Acknowledgment}
\input{./chapter/acknowledgement.tex}

%
%
%
%
\printbibliography

\end{document}

%% file: chapter/acronyms.tex
\newacro{5g}[5G]{fifth generation}
\newacro{6g}[6G]{sixth generation}
\newacro{as}[AS]{Authentication Server}
\newacro{atm}[ATM]{Automated Teller Machine}
\newacro{ai}[AI]{artificial intelligence}
\newacroplural{ais}[AIs]{artificial intelligences}
\newacro{auc}[AUC]{area under the curve}

\newacro{b5g}[B5G]{Beyond 5G}
\newacro{ban}[BAN]{Body Area Network}
\newacro{bsi}[\textit{BSI}]{\textit{Federal Office for Information Security}}
\newacro{bdr}[BDR]{Bit Disagreement Rate}
\newacro{bs}[BS]{Base Station}
\newacro{bilstm}[BiLSTM]{bidirectional long short-term memory}
\newacro{ca}[CA]{Certification Authority}
\newacro{cav}[CAV]{Connected Autonomous Vehicles}
\newacro{cc}[CC]{Common Criteria}
\newacro{cir}[CIR]{Channel Impulse Response}
\newacro{cr}[CR]{Challenge-Response}
\newacro{cpu}[CPU]{Central Processing Unit}
\newacro{cpps}[CPPS]{Cyber-Physical Production System}
\newacro{crl}[CRL]{Certificate Revocation List}
\newacro{csi}[CSI]{Channel State Information}
\newacro{crke}[CRKE]{Channel-Reciprocity Based Key Extraction}
\newacro{ctf}[CTF]{Channel Transfer Function}
\newacro{cotf}[COTF]{Commercial-off-the-Shelf}
\newacro{cmos}[CMOS]{Complementary Metal-Oxide-Semiconductors}
\newacro{cnn}[CNN]{Convolutional neural network}
\newacro{cidds}[CIDDS]{Coburg Intrusion Detection Datasets}

\newacro{dos}[DoS]{denial-of-service}
\newacro{ddos}[DDoS]{distributed-denial-of-service}
\newacro{dna}[DNA]{Deoxyribonucleic Acid}
\newacro{dtls}[DTLS]{Datagram Transport Layer Security}
\newacro{dct}[DCT]{Discrete Cosine Transformation}
\newacro{dlt}[DLT]{Distributed Ledger Technology}
\newacro{dl}[DL]{deep learning}
\newacro{eal}[EAL]{Evaluation Assurance Level}
\newacro{ecc}[ECC]{Elliptic Curve Cryptography}
\newacro{ecg}[ECG]{Electrocardiogram}
\newacro{eda}[EDA]{exploratory data analysis}
\newacro{eeg}[EEG]{Electroencephalogram}
\newacro{embb}[eMBB]{enhanced Mobile Broadband}
\newacro{emg}[EMG]{Electromyogram}
\newacro{eog}[EOG]{Electrooculography}
\newacro{enb}[eNodeB]{Evolved Node B}
\newacro{er}[ER]{Extended Reality}
\newacro{ema}[EMA]{exponential moving average}
\newacro{fpga}[FPGA]{Field Programmable Gate Array}
\newacro{fdd}[FDD]{Frequency Division Duplexing}
\newacro{fpr}[FPR]{false positive rates}
\newacro{gdpr}[GDPR]{General Data Protection Regulation}
\newacro{gd}[G\&D]{Giesecke \& Devrient}
\newacro{gcn}[GCN]{graph convolutional network}
\newacro{gnn}[GNN]{graph neural network}

\newacro{h2m}[H2M]{Human-to-Machine}
\newacro{h2s}[H2S]{Human-to-Service}
\newacro{hmac}[HMAC]{Keyed-Hash Message Authentication Code}
\newacro{htc}[HTC]{Hologaphic-Type Communication}
\newacro{hotp}[HOTP]{HMAC-based One-time Password Algorithm}
\newacro{hsm}[HSM]{Hardware Security Module}
\newacro{ics}[ICS]{Industrial Control System}
\newacro{iacs}[IACS]{Industrial Automation and Control System}
\newacro{ioe}[IoE]{Internet of Everything}
\newacro{iiot}[IIoT]{Industrial Internet of Things}
\newacro{iot}[IoT]{Internet of Things}
\newacro{io}[I/O]{Input/Output}
\newacro{ic}[IC]{Integrated Circuit}
\newacro{id}[ID]{Identifier}
\newacro{ids}[IDS]{intrusion detection system}
\newacro{irs}[IRS]{Intelligent Reflecting Surface}
\newacro{istn}[ISTN]{Integrated Space and Terrestrial Network}
\newacro{it}[IT]{Information Technology}
\newacro{itu}[ITU]{International Telecommunication Union}
\newacro{jcop}[JCOP]{Java Card Open Platform}
\newacro{kba}[KBA]{Knowledge Based Authentication}
\newacro{kdf}[KDF]{Key Derivation Function}
\newacro{knn}[kNN]{k-Nearest Neighbors}

\newacro{led}[LED]{Light Emitting  Diode}
\newacro{lte}[LTE]{Long Term Evolution}
\newacro{ltea}[LTE-A]{Long Term Evolution Advanced}
\newacro{lr}[LR]{Linear Regression}
\newacro{los}[LoS]{Line of Sight}
\newacro{lorawan}[LoRaWAN]{Long Range Wide Area Network}
\newacro{lstm}[LSTM]{long short-term memory}
\newacro{mbb}[MBB]{Mobile Broadband}
\newacro{mfa}[MFA]{Multi-Factor Authentication}
\newacro{mcc}[MCC]{Mobile Cloud Computing}
\newacroplural{mcus}[MCUs]{Microcontroller Units}
\newacro{m2m}[M2M]{Machine-to-Machine}
\newacro{m2s}[M2S]{Machine-to-Service}
\newacro{mimo}[MIMO]{Multiple Input Multiple Output}
\newacro{mmimo}[mMIMO]{massive Multiple Input Multiple Output}
\newacro{ml}[ML]{machine learning}
\newacro{mulc}[mULC]{massive Ultra-Reliable Low-Latency Communication}
\newacro{mmtc}[MMTC]{massive Machine Type Communication}
\newacro{mmg}[MMG]{Mechanomyogram}
\newacro{multos}[MULTOS]{Multi-Application Smart Card Operating System}
\newacro{mux}[MUX]{Multiplexer}
\newacro{mnc}[MNC]{Mobile Network Code}
\newacro{me}[ME]{Mobile Environment}
\newacro{mac}[MACs]{Message Authentication Codes}
\newacro{mse}[MSE]{mean squared error}
\newacro{mcc}[MCC]{Matthews Correlation Coefficient}
\newacro{ngmn}[NGMN]{Next Generation Mobile Networks}
\newacro{nic}[NIC]{Network Interface Controller}
\newacro{nist}[NIST]{National Institute of Standards and Technology}
\newacro{nn}[NN]{neural network}
\newacro{nids}[NIDS]{Network Intrusion Detection Systems}
\newacro{oauth}[OAuth]{Open Authentication}
\newacro{ocra}[OCRA]{\ac{oauth} Challenge-Response Algorithm}
\newacro{ocsp}[OCSP]{Online Certificate Status Protocol}
\newacro{otp}[OTP]{One-Time Password}
\newacro{pap}[PAP]{Password-Authentication-Protocol}
\newacro{physec}[PhySec]{Physical Layer Security}
\newacro{pfs}[PFS]{Perfect Forward Secrecy}
\newacro{pin}[PIN]{Personal Identification Number}
\newacro{pkc}[PKC]{Public Key Cryptography}
\newacro{pki}[PKI]{Public Key Infrastructure}
\newacro{ppg}[PPG]{Photoplethysmography}
\newacro{prng}[PRNG]{Pseudo Random Number Generator}
\newacro{puf}[PUF]{Physically Unclonable Function}
\newacroplural{pufs}[PUFs]{Physically Unclonable Functions}
\newacro{pla}[PLA]{Physical Layer Authentication}
\newacro{qr}[QR]{Quick Response}
\newacro{qos}[QoS]{quality of service}
\newacro{rat}[RAT]{Radio Access Technology}
\newacro{radius}[RADIUS]{Remote Authentication Dial-In User Service}
\newacro{ram}[RAM]{Random-Access Memory}
\newacro{ran}[RAN]{Radio Access Networks}
\newacro{rf}[RF]{random forest}
\newacro{rfid}[RFID]{Radio-Frequency Identification}
\newacro{ris}[RIS]{Reconfigurable Intelligent Surface}
\newacro{rng}[RNG]{Random Number Generator}
\newacro{ro}[RO]{Ring-Oscillator}
\newacro{rom}[ROM]{Read-Only Memory}
\newacro{rs}[RS]{Reed-Solomon}
\newacro{rsa}[RSA]{Rivest-Shamir-Adleman}
\newacro{rssi}[RSSI]{Received Signal Strength Indicator}
\newacro{rsrp}[RSRP]{Reference Signal Received Power}
\newacro{re}[RE]{Resource Elements}
\newacro{rnn}[RNN]{recurrent neural network}
\newacro{roc}[ROC]{receiver operating characteristic curve}

\newacro{sdn}[SDN]{software-defined networking}
\newacro{sdr}[SDR]{Software-Defined Radio}
\newacro{seccos}[SECCOS]{Secure Chip Card Operating System}
\newacro{sip}[SIP]{Session Initiation Protocol}
\newacro{skg}[SKG]{Secret Key Generation}
\newacro{sram}[SRAM]{Static Random Access Memory}
\newacro{srs}[SRS]{Software Radio Systems}
\newacro{starcos}[STARCOS]{Smart Card Chip Operating System}
\newacro{sha}[SHA]{Secure Hash Algorithm}
\newacro{se}[SE]{Static Environment}
\newacro{svm}[SVM]{support vector machine}
\newacro{ssl}[SSL]{semi-supervised learning}
\newacro{tcg}[TCG]{Trusted Computing Group}
\newacro{tpm}[TPM]{Trusted Platform Module}
\newacro{tls}[TLS]{Transport Layer Security}
\newacro{trng}[TRNG]{True Random Number Generator}
\newacro{tsn}[TSN]{Time-Sensitive Networking}
\newacro{tofu}[TOFU]{Trust On First Use}
\newacro{tufu}[TUFU]{Trust Upon First Use}
\newacro{totp}[TOTP]{Time-based One-time Password Algorithm}
\newacro{tpr}[TPR]{true positive rates}
\newacro{uav}[UAV]{Unmanned Aerial Vehicles}
\newacro{usb}[USB]{Universal Serial Bus}
\newacro{usrp}[USRP]{Universal Software Radio Peripheral}
\newacro{uhd}[UHD]{USRP Hardware Driver}
\newacro{usim}[USIM]{Universal Subscriber Identity Module}
\newacro{ue}[UE]{User Equipment}
\newacro{urllc}[URLLC]{Ultra-Reliable Low-Latency Communication}
\newacro{ulbc}[ULBC]{Ultra-Reliable Low-Latency Broadband Communication}
\newacro{umbb}[uMBB]{ubiquitous Mobile Broadband}  
\newacro{ummimo}[UM-MIMO]{Ultra-Massive MIMO}
\newacro{vlc}[VLC]{Visible Light Communication}
\newacro{warp}[WARP]{Wireless open-Access Research Platform}

%% file: chapter/introduction.tex
\label{sec:introduction}
\subsection{Background \& Motivation}
Cloud computing has become a core component of modern digital infrastructure, supporting critical services across enterprise, industrial, and public sectors. The scale, elasticity, and heterogeneity of cloud environments generate large volumes of high-dimensional network traffic, rendering traditional signature-based security mechanisms increasingly ineffective \cite{Buczak2016}. As a result, \ac{ml}–based \ac{nids} have emerged as a central element of cloud security architectures, enabling automated detection of malicious activity from network traffic patterns \cite{Sommer2010}.

Despite their promise, \ac{ml}-driven \ac{nids} face significant challenges in real-world cloud deployments \cite{zuech2015intrusion}. High quality labeled attack data is scarce due to the cost of manual annotation, privacy constraints, and the continuous emergence of novel threats. Moreover, cloud network traffic is inherently non-stationary, meaning they have variable scaling, workload dynamics, and contain evolving user behavior, which induce persistent changes in traffic distributions. These challenges are compounded by adaptive adversaries who deliberately modify their behavior to evade learning-based detection systems.

\subsection{Limitations of Existing Network Intrusion Detection Approaches}
Early learning-based \ac{nids} relied on supervised learning, achieving strong performance on benchmark datasets but limited generalization in real-world deployments \cite{ring2019survey}. These approaches assume the availability of comprehensive labeled data, which is rarely feasible. To address this, recent work has explored unsupervised and \ac{ssl} methods that leverage large volumes of unlabeled traffic \cite{ferrag2019deep}.

However, most semi-supervised \ac{nids} assume unlabeled data is predominantly benign. In cloud environments, this assumption often fails, as unlabeled traffic may contain unknown attacks, low-and-slow exfiltration, or adversarially crafted flows that corrupt learning. As a result, such models are prone to performance degradation, false negatives, and concept drift \cite{Sommer2010}. Concurrently, advances in adversarial \ac{ai} show that attackers can exploit \ac{ml}-based defenses by mimicking benign patterns or manipulating unlabeled data, raising concerns about the robustness of \ac{ai}-driven \ac{nids} \cite{shone2018deep}.

\subsection{Research Gap and Key Insight}
While \ac{ssl} offers a promising approach under label scarcity, existing methods largely overlook the adversarial and non-stationary nature of cloud network traffic \cite{lian2025semi}. In particular, current semi-supervised \ac{nids} lack mechanisms to (i) suppress malicious samples in unlabeled data, (ii) remain robust under temporal and distributional drift, and (iii) generalize across heterogeneous cloud environments.

Our key insight is that cloud traffic exhibits strong temporal structure that can be exploited even with limited labels. Benign traffic tends to be temporally stable, whereas malicious activity introduces subtle but meaningful temporal inconsistencies over time. By explicitly leveraging temporal agreement and incorporating robustness-aware objectives, unlabeled data can be utilized effectively while limiting adversarial contamination.

\subsection{Paper Contributions}
To address these challenges, this paper makes the following contributions:
\begin{enumerate}
    \item \textbf{Problem Formulation:} We formalize the problem of robust semi-supervised network intrusion detection in cloud environments, explicitly accounting for label scarcity, adversarial contamination, and non-stationary traffic distributions.
    
    \item \textbf{Methodology:} We propose \textbf{RSST-NIDS}, a robust semi-supervised temporal intrusion detection framework that combines supervised learning with consistency regularization, confidence-aware pseudo-labeling, and selectively applied temporal invariance to mitigate the influence of malicious unlabeled data.
    
    \item \textbf{Evaluation:} We conduct a comprehensive evaluation on publicly available, real-world network traffic datasets under limited label and cross-dataset settings, demonstrating improved detection performance, robustness to distribution shift, and label efficiency compared to state-of-the-art approaches.
    
    \item \textbf{Security Implications:} We analyze the implications of deploying semi-supervised learning in adversarial cloud environments and show how robustness-aware temporal learning can strengthen \ac{ai}-based network defenses.
\end{enumerate}

%% file: chapter/works.tex
\label{sec:works}
\subsection{Network Intrusion Detection in Cloud Environments}
The deployment of \ac{nids} in cloud environments introduces challenges beyond traditional networks, including large-scale heterogeneity, elastic workloads, and evolving threat surfaces. Domain-invariant approaches such as DI-NIDS \cite{layeghy2023di} address this by learning features invariant across domains via adversarial training, achieving strong performance when target-domain data is available.

However, this assumption is often unrealistic in cloud settings, where future traffic domains and labels are unavailable. Moreover, enforcing domain invariance may suppress cues indicative of emerging or stealthy attacks. In contrast, we address a complementary setting focused on \emph{domain-agnostic robustness} under label scarcity and adversarial contamination, without requiring target-domain access. Instead of explicit domain alignment, we leverage temporal self-consistency and conservative use of unlabeled data to improve generalization across heterogeneous environments.

\subsection{Deep Learning for Network Traffic Analysis}
Deep learning has become the dominant paradigm for network traffic analysis, enabling automated feature learning from flow-level data. \acp{cnn} capture spatial correlations \cite{chen2020novel}, while \acp{rnn} and \ac{lstm} model temporal dependencies \cite{park2020rnn}. More recently, Transformer-based models have improved the modeling of long-range temporal structure \cite{akuthota2025transformer}.

However, most deep learning–based \ac{nids} are trained in fully supervised settings and evaluated within static datasets. As a result, performance degrades under distribution shift, temporal drift, and class imbalance typical of real cloud traffic. Moreover, reliance on curated datasets with clean labels limits their applicability in adversarial and non-stationary environments, motivating approaches that leverage unlabeled data while maintaining robustness.

\subsection{Unsupervised and Semi-Supervised Learning for NIDS}
Unsupervised methods typically model normal traffic and detect anomalies via deviation, with autoencoders widely used for reconstruction-based detection. While reducing reliance on labels, these approaches often suffer from high false positives and difficulty distinguishing benign anomalies from attacks \cite{choi2019unsupervised}.

Semi-supervised methods address this by combining limited labeled data with large unlabeled sets. Approaches such as AnomalyAID \cite{yuan2024anomalyaid} use reconstruction-based pseudo-label verification to filter noisy samples, but may be less sensitive to temporally evolving attacks. Similarly, SS-VTCN \cite{jia2022semi} integrates generative temporal reconstruction with discriminative learning, capturing short-term dependencies but potentially enforcing invariance on slowly evolving attacks.

In contrast, our approach selectively applies temporal invariance only to benign-consistent windows, avoiding global reconstruction constraints and reducing the risk of suppressing attack-specific temporal patterns.

\subsection{Adversarial Machine Learning in Network Security}
The vulnerability of machine learning models to adversarial manipulation has been extensively studied in other domains. However, its implications for network intrusion detection remain less explored. Existing adversarial defense techniques often assume fully labeled datasets or rely on adversarial training strategies that are difficult to apply in semi-supervised settings. Moreover, many defenses focus on static feature perturbations rather than adaptive, temporally structured attacks commonly observed in network traffic \cite{shen2025robust}.

In contrast to prior work, this paper explicitly considers the interaction between adversarial behavior, unlabeled data contamination, and temporal drift in cloud network intrusion detection. Rather than assuming benign unlabeled traffic, we adopt confidence-aware and temporality-driven learning objectives that conservatively exploit unlabeled data while suppressing the influence of uncertain or malicious samples. By leveraging temporal consistency as a robustness signal and constraining invariance enforcement to benign-consistent windows, the proposed framework addresses key limitations of existing semi-supervised and adversarially robust \ac{nids}.
\\
\\
To the best of our knowledge, this work represents one of the first systematic investigations of robust semi-supervised learning for cloud network intrusion detection under realistic, adversarial conditions using publicly available traffic datasets. The proposed approach complements existing deep learning based \ac{nids} by providing a security-aware framework that challenges that are suitable to practical cloud deployments and explicitly accounts for label scarcity, temporal non-stationarity, and adversarial contamination.

%% file: chapter/method.tex
\label{sec:method}

\subsection{Problem Setting and Objective}
We consider a cloud network intrusion detection scenario in which traffic is observed as a continuous stream of flow-level measurements. Let $\mathcal{X} = \{x_t\}_{t=1}^{N}$ denote a sequence of traffic windows, where each sample $x_t \in \mathbb{R}^{F \times T}$ represents $F$ flow-level features aggregated over a temporal window of length $T$.

We adopt a semi-supervised learning setting comprising a small labeled subset $\mathcal{X}_L = \{(x_i, y_i)\}_{i=1}^{N_L}$ with labels $y_i \in \{0,1\}$ (benign or malicious), and a large unlabeled subset $\mathcal{X}_U = \{x_j\}_{j=1}^{N_U}$, where $N_U \gg N_L$. Importantly, $\mathcal{X}_U$ may contain unknown or adversarial traffic, and the underlying data distribution is non-stationary due to workload variation and attacker adaptation.

The objective is to learn a detection function:
\[
f_\theta : \mathbb{R}^{F \times T} \rightarrow [0,1],
\]
which predicts the probability of malicious activity while (i) exploiting unlabeled data efficiently, (ii) remaining robust to adversarial contamination, and (iii) maintaining stable performance under temporal drift.

\subsection{Threat Model and Design Assumptions}
We adopt a black-box adversarial threat model consistent with practical cloud network settings. The adversary has no access to model parameters or gradients, but can manipulate network traffic subject to protocol and feature validity constraints.
Two attack capabilities are considered. \emph{Poisoning} attacks inject malicious flows into the unlabeled training stream at contamination rates of 5-15\%, crafting samples that resemble benign traffic in basic statistics while preserving attack semantics. \emph{Evasion} attacks are generated via constrained feature blending followed by projection onto the valid problem space: features are clipped and normalized to satisfy NetFlow constraints (e.g., non-negativity, bounded rates, and consistency between packet counts and byte volumes), and invalid samples are discarded.

White-box attacks are out of scope, as they are less realistic at the network level. While an adaptive adversary could attempt to craft high confidence, slowly evolving malicious traffic, RSST-NIDS does not claim worst case robustness; instead, it aims to raise the cost of evasion by requiring sustained temporal stability and student-teacher agreement under realistic constraints.

\subsection{Traffic Representation and Temporal Modeling}
To ensure scalability and privacy preservation, the proposed framework operates exclusively on flow-level features, including packet and byte counts, flow duration, inter-arrival time statistics, and header-level entropy measures. Flows are aggregated into sliding temporal windows
\[
x_t = \{f_{t-T+1}, \ldots, f_t\},
\]
which enables modeling of burst-based attacks, low-and-slow intrusions, and periodic malicious behavior commonly observed in cloud environments.

Temporal dependencies are captured using a Transformer-based encoder that produces a latent representation $z_t$ for each window. Transformers are selected for their ability to model long range temporal dependencies. However, the proposed learning framework is architecture-agnostic and can be instantiated with lighter temporal encoders (e.g., TCNs or LSTMs) when computational constraints dominate.

\subsection{Robust Semi-Supervised Learning Framework}
RSST-NIDS combines supervised learning on labeled data with conservative semi-supervised learning on unlabeled traffic. The core design principle is to exploit unlabeled data only when there is sufficient evidence of reliability, thereby limiting the impact of adversarial contamination and noisy pseudo-labels.

Training integrates four complementary components: supervised loss on labeled samples ($\mathcal{L}_{\text{sup}}$), consistency regularization under traffic-preserving perturbations ($\mathcal{L}_{\text{cons}}$), confidence-aware pseudo-labeling for reliable unlabeled data ($\mathcal{L}_{\text{pseudo}}$), and selectively applied temporal regularization ($\mathcal{L}_{\text{temp}}$). The overall objective ($\mathcal{L}_{\text{total}}$) is
\[
\mathcal{L}_{\text{total}} = \mathcal{L}_{\text{sup}} + \lambda_c \mathcal{L}_{\text{cons}}
+ \lambda_p \mathcal{L}_{\text{pseudo}} + \lambda_t \mathcal{L}_{\text{temp}},
\]

Labeled samples are optimized using standard cross-entropy loss. For unlabeled samples, consistency regularization enforces stable predictions under carefully constrained augmentations that preserve NetFlow semantics and protocol validity. Temporal jitter is implemented by shifting window boundaries by at most $\pm5\%$ of the window length, without reordering or duplicating flows, thereby preserving causal ordering. Feature masking is restricted to non-causal, derived features (e.g., entropy or aggregate statistics), while core counters such as packet counts, byte volumes, and flow durations are never masked; masked features are replaced with dataset level means, to avoid implausible artifacts. Noise injection is applied as small Gaussian perturbations bounded by feature specific variance and clipped to maintain non-negativity and protocol consistent ranges. Augmented samples violating NetFlow constraints (e.g., packet counts $\geq 1$, byte counts consistent with packet counts, non-negative durations) are discarded.

An \ac{ema} teacher model $f_{\theta'}$ is maintained to stabilize predictions on unlabeled data and to support agreement based filtering. The teacher is initialized from the student and updated as:
\[
\theta' \leftarrow \alpha \theta' + (1-\alpha)\theta,
\]
with decay $\alpha = 0.99$. Importantly, the \ac{ema} teacher is not used to generate pseudo-labels or enforce global consistency. Instead, it serves solely as a conservative reference for selective temporal regularization, ensuring that temporal invariance is never imposed on uncertain or potentially malicious traffic.

\subsection{Selective Temporal Robustness}
Temporal invariance is enforced selectively to avoid suppressing attack specific dynamics. An unlabeled window $x_t$ is deemed \emph{benign-consistent} only if it satisfies all of the following criteria:

\textbf{(i) Confidence:} $\max(\hat{y}_t, 1-\hat{y}_t) \ge \tau_b$

\textbf{(ii) Student-Teacher Agreement:}
\[
|f_\theta(x_t) - f_{\theta'}(x_t)| \le \delta_{\text{EMA}},
\]

\textbf{(iii) Temporal Stability:} Temporal stability is assessed over a short neighborhood of adjacent windows to identify persistent benign behavior. During training and offline evaluation, stability is computed over a symmetric window $\{x_{t-k}, \ldots, x_{t+k}\}$ to improve robustness against transient fluctuations. In deployment, where future windows are unavailable, the same criterion is applied using a causal variant that considers only past windows $\{x_{t-k}, \ldots, x_t\}$ with a bounded decision delay of $k$ windows.
This design ensures that RSST-NIDS remains deployable in online settings while allowing more reliable stability estimation during training. All reported online-relevant results use the causal formulation.
\[
\mathrm{Var}\left(\{f_\theta(x_{t-k}), \ldots, f_\theta(x_{t+k})\}\right) \le \delta_{\text{temp}}.
\]

Only windows satisfying all three conditions are included in the benign-consistent set $\mathcal{B}$. For windows admitted to the benign-consistent set $\mathcal{B}$, we enforce temporal invariance directly in the latent representation space produced by the temporal encoder. Specifically, let $z_t = E_\theta(x_t)$ denote the latent embedding of window $x_t$. The temporal invariance loss is defined as:
\[
\mathcal{L}_{\text{temp}} =
\frac{1}{|\mathcal{B}|}
\sum_{x_t \in \mathcal{B}}
\left\| z_t - z_{t+\Delta} \right\|_2^2,
\]
where $\Delta$ denotes a fixed temporal offset corresponding to adjacent windows (i.e., $\Delta=1$ in all experiments).

This loss penalizes abrupt changes in latent representations across time for traffic deemed benign-consistent, encouraging local temporal smoothness while avoiding constraints on uncertain or potentially malicious windows. We operate in the latent space rather than on logits or probabilities to regularize temporal representations without directly biasing the decision boundary.

\subsection{Training Procedure and Parameterization}
Algorithm~\ref{alg:ssl_nids} summarizes the complete training procedure. Labeled samples contribute supervised gradients, while unlabeled samples are incorporated conservatively through consistency regularization, confidence-aware pseudo-labeling, and selective temporal regularization. This design ensures that unlabeled data improves representation learning without amplifying noise or adversarial influence.

Unless stated otherwise, gating parameters are fixed across all experiments: $\tau_b = 0.95$, $\delta_{\text{EMA}} = 0.05$, temporal neighborhood size $k=2$, and temporal stability threshold $\delta_{\text{temp}} = 0.01$. Temporal invariance is applied between adjacent windows ($\Delta=1$). These values are selected via validation on CIC-IDS2017 and reused unchanged for all cross-dataset evaluations, demonstrating that RSST-NIDS does not require dataset specific retuning.

\FloatBarrier

\begin{algorithm}[t]
\small
\caption{Semi-Supervised Intrusion Detection Training}
\label{alg:ssl_nids}
\begin{algorithmic}[1]
\REQUIRE Labeled dataset $\mathcal{X}_L = \{(x_i, y_i)\}_{i=1}^{N_L}$, unlabeled dataset $\mathcal{X}_U = \{x_j\}_{j=1}^{N_U}$, temporal encoder $E_\theta$, classifier $C_\phi$, confidence threshold $\tau$, loss weights $\lambda_c, \lambda_p, \lambda_t$, number of epochs $E$

\ENSURE Trained intrusion detection model $f(x) = C_\phi(E_\theta(x))$

\STATE Initialize $\theta, \phi$ randomly
\FOR{epoch = 1 to $E$}

    \STATE \textbf{Supervised Learning Phase:}
    \STATE Sample mini-batch $B_L \subseteq \mathcal{X}_L$
    \FOR{each $(x_i, y_i) \in B_L$}
        \STATE $z_i \gets E_\theta(x_i)$
        \STATE $\hat{y}_i \gets C_\phi(z_i)$
    \ENDFOR
    \STATE Compute supervised loss: $\mathcal{L}_{\text{sup}} \gets \text{CrossEntropy}(\hat{y}_i, y_i)$

    \STATE \textbf{Unlabeled Data Processing:}
    \STATE Sample mini-batch $B_U \subseteq \mathcal{X}_U$
    \STATE Initialize confident set $S \gets \emptyset$
    \FOR{each $x_j \in B_U$}
        \STATE Generate perturbed view: $x_j' \gets \mathcal{A}(x_j)$ \COMMENT{temporal jitter, masking, noise}
        \STATE Forward pass: $z_j \gets E_\theta(x_j),\ z_j' \gets E_\theta(x_j')$
        \STATE Predictions: $\hat{y}_j \gets C_\phi(z_j),\ \hat{y}_j' \gets C_\phi(z_j')$
        \STATE Consistency loss: $\mathcal{L}_{\text{cons}} += \|\hat{y}_j - \hat{y}_j'\|_2^2$
        \STATE Confidence-aware pseudo-labeling:
        \IF{$\hat{y}_j \ge \tau$}
            \STATE $\tilde{y}_j \gets 1$
            \STATE $S \gets S \cup \{(x_j, \tilde{y}_j)\}$
        \ELSIF{$\hat{y}_j \le 1 - \tau$}
            \STATE $\tilde{y}_j \gets 0$
            \STATE $S \gets S \cup \{(x_j, \tilde{y}_j)\}$
        \ENDIF
    \ENDFOR
    \STATE \textbf{Benign-Consistent Window Selection:}
    \FOR{each $x_t \in B_U$}
        \IF{confidence, EMA agreement, and temporal stability criteria satisfied}
            \STATE $\mathcal{B} \gets \mathcal{B} \cup \{x_t\}$
        \ENDIF
    \ENDFOR

\STATE \textbf{Temporal Invariance Loss:}
\FOR{each $x_t \in \mathcal{B}$}
    \STATE $\mathcal{L}_{\text{temp}} += \|E_\theta(x_t) - E_\theta(x_{t+\Delta})\|_2^2$
\ENDFOR

    \STATE \textbf{Pseudo-Label Supervision:}
    \IF{$S \neq \emptyset$}
        \FOR{each $(x_j, \tilde{y}_j) \in S$}
            \STATE $\hat{y}_j \gets C_\phi(E_\theta(x_j))$
            \STATE $\mathcal{L}_{\text{pseudo}} += \text{CrossEntropy}(\hat{y}_j, \tilde{y}_j)$
        \ENDFOR
    \ELSE
        \STATE $\mathcal{L}_{\text{pseudo}} \gets 0$
    \ENDIF

    \STATE \textbf{Total Loss:} 
    \STATE $\mathcal{L}_{\text{total}} \gets \mathcal{L}_{\text{sup}} + \lambda_c \mathcal{L}_{\text{cons}} + \lambda_p \mathcal{L}_{\text{pseudo}} + \lambda_t \mathcal{L}_{\text{temp}}$
    \STATE Update $\theta, \phi$ using gradient descent on $\mathcal{L}_{\text{total}}$

\ENDFOR

\RETURN $f(x)$
\end{algorithmic}
\end{algorithm}

%% file: chapter/setup.tex
\label{sec:setup}
\subsection{Datasets and Evaluation Scope}

We evaluate RSST-NIDS on three publicly available flow-based NIDS datasets that capture realistic, heterogeneous cloud traffic. CIC-IDS2017 serves as the primary training and in-domain evaluation dataset, containing diverse benign and malicious activities (e.g., brute force, DoS/DDoS, botnet, infiltration). To assess robustness under distribution shift, we further evaluate on CSE-CIC-IDS2018, which extends CIC-IDS2017 with larger traffic volumes and additional attack types, and UNSW-NB15, which includes modern attack families and realistic benign traffic. All datasets are converted into flow-level windows before training. Table~\ref{tab:data_stats} reports the number of windows after cleaning and preprocessing; natural class imbalance is preserved to reflect realistic cloud environments.

\begin{table}[ht]
\centering
\caption{Dataset Statistics After Cleaning and Windowing}
\label{tab:data_stats}
\begin{tabular}{lccc}
\hline
\textbf{Dataset} & \textbf{Split} & \textbf{Samples} & \textbf{Attack Ratio (\%)} \\
\hline
\multirow{3}{*}{CIC-IDS2017} &
Train & 180K & 18.5 \\
& Val   & 40K  & 17.9 \\
& Test  & 60K  & 18.2 \\
\hline
CSE-CIC-IDS2018 & Test  & 120K & 16.7 \\
\hline
UNSW-NB15 & Test  & 100K & 14.2 \\
\hline
\end{tabular}
\end{table}

Following established guidelines on \ac{nids} dataset reliability and feature leakage, we remove (i) flows with missing or undefined features, (ii) duplicate flow records, (iii) flows with inconsistent NetFlow statistics (e.g., byte count $<$ packet count), and (iv) mis-labeled or anomalous CIC-IDS2017 segments. All identifier and leakage-prone fields (timestamps, FlowIDs, IPs, ports, dataset-specific flags) are excluded to prevent artifact-based learning and ensure cross-dataset comparability.
\\
Consistent with prior analyses, we omit CIC-IDS2017 regions exhibiting labeling inconsistencies or abnormal traffic distributions (e.g., portions of WebAttack and Infiltration days). Filtering is applied conservatively to avoid skewing class ratios, and the same rules are used across datasets.
\\
CIC-IDS2017 is split strictly by time: earlier days for training, intermediate traffic for validation, and later days for testing, with no day shared across splits. Cross-dataset experiments train solely on CIC-IDS2017 and evaluate on CSE-CIC-IDS2018 and UNSW-NB15 without fine-tuning.
\\
All metrics are averaged over $n=5$ runs with different random seeds. Statistical significance is assessed using paired Wilcoxon signed-rank tests ($p < 0.01$), where pairs correspond to identical seeds and splits across models.

\subsection{Feature Extraction and Data Sanitization}
To ensure consistency across datasets, we extract flow-level features using CICFlowMeter v4, avoiding packet payloads to preserve privacy and scalability. Fields known to introduce information leakage or dataset-specific artifacts e.g., timestamps, flow identifiers, IP addresses, labels, and exporter-specific flags are removed prior to training. Traffic segments previously identified as mislabeled or anomalous are excluded following established dataset cleaning practices. This preprocessing ensures that performance gains reflect model robustness rather than unintended leakage. All augmented samples used for consistency regularization are validated against the same NetFlow constraints applied during preprocessing, ensuring protocol-consistent traffic throughout training.

\subsection{Training Protocol}
Training follows a strictly time-ordered protocol to preserve causality and emulate realistic cloud traffic observation. Models are warm-started on labeled data, after which semi-supervised learning is enabled by progressively incorporating unlabeled samples. Pseudo-labeling uses a conservative confidence threshold $\tau$, and samples failing confidence or agreement criteria are excluded from supervision. Unless stated otherwise, experiments use temporal windows of length $T=20$ with stride 5 and $F=78$ flow-level features. Mini-batches contain labeled and unlabeled samples in a 1:4 ratio (batch size 256). The Transformer encoder has 4 layers, 4 attention heads, hidden size 128, and dropout 0.1. Traffic-preserving augmentations include temporal jitter ($\pm5\%$), feature masking (10\%), and Gaussian noise ($\sigma=0.01$). Loss weights are $\lambda_c=1.0$, $\lambda_p=0.5$, and $\lambda_t=0.2$. Models are trained for 50 epochs using Adam (learning rate $10^{-3}$, cosine decay). In addition to the default 10\% labeled setting, we evaluate a 5\% labeled regime to assess label efficiency. To prevent temporal leakage, datasets are split chronologically. For CIC-IDS2017, training uses Monday and Tuesday traffic, validation uses Wednesday, and testing uses Thursday and Friday, ensuring no overlap across splits. Cross-dataset experiments are trained exclusively on CIC-IDS2017 and evaluated on CSE-CIC-IDS2018 or UNSW-NB15 without fine-tuning. This setup ensures that reported results reflect generalization under temporal drift rather than memorization.

\subsection{Baseline Methods}
We compare RSST-NIDS against representative supervised, unsupervised, and semi-supervised NIDS baselines under identical feature representations, labeled data budgets, splits, and training settings to ensure fairness.
Supervised baselines include CNN-, LSTM-, and Transformer-based models, covering common spatial and temporal architectures. The unsupervised baseline is an autoencoder trained on benign traffic, where anomalies are detected via reconstruction error. For semi-supervised learning, we consider four approaches: (i) FixMatch-style pseudo-labeling \cite{sohn2020fixmatch} with high-confidence hard labels, (ii) SimCLR-style contrastive pretraining \cite{kim2020adversarial} followed by fine-tuning, (iii) class-aware SSL (CCSSL) \cite{kim2023sound} combining confidence-weighted pseudo-labels with contrastive learning, and (iv) Mean Teacher \cite{tarvainen2017mean}, enforcing consistency under perturbations without temporal gating. None explicitly model temporal consistency or perform benign-consistency verification.

We exclude AnomalyAID, SS-VTCN, and DI-NIDS due to differing assumptions: reconstruction-based verification, global temporal reconstruction, or access to target-domain data during training. In contrast, RSST-NIDS targets conservative semi-supervised learning under adversarial contamination without target-domain access, making the selected baselines the most relevant for comparison. 

\subsection{Adversarial Contamination and Evasion Evaluation}

We evaluate robustness under realistic black-box adversarial settings constrained by protocol-valid traffic.

\textbf{Poisoning.} Adversarial samples are injected into the unlabeled set at 5\%, 10\%, and 15\% contamination. Malicious flows are modified to match benign first-order statistics (e.g., duration, packet count, byte rate) while preserving temporal structure and attack semantics.

\textbf{Evasion.} At test time, adversarial flows are generated via constrained traffic morphing:
\begin{align}
x_{\text{adv}} = \alpha x_{\text{attack}} + (1-\alpha)x_{\text{benign}}, \quad \alpha \in [0.6, 0.8]
\end{align}
All samples are projected onto valid NetFlow feature space. We also evaluate transfer-based attacks, where morphing parameters are tuned on a surrogate model and applied to RSST-NIDS.

These attacks capture realistic constraints faced by network-level adversaries and provide a conservative assessment of robustness under black-box conditions.

%% file: chapter/results.tex
\label{sec:results}
This section evaluates the proposed RSST-NIDS framework under realistic cloud deployment conditions. We first assess detection performance and generalization under limited label and cross-dataset settings. We then examine robustness to adversarial contamination and evasion, followed by ablation, sensitivity, and calibration analyses.

\subsection{Overall Detection Performance and Generalization}

We begin by comparing RSST-NIDS with supervised, unsupervised, and semi-supervised baselines when only 10\% of the training data is labeled, with the remaining traffic treated as unlabeled and potentially adversarial. Table~\ref{tab:model_comparison} summarizes in-domain performance on CIC-IDS2017. RSST-NIDS consistently outperforms all baselines across accuracy, F1-score, and AUROC, achieving a 2-3\% absolute F1-score improvement over recent \ac{ssl}-based \ac{nids} despite severe class imbalance and limited labeled data. These gains indicate that conservative exploitation of unlabeled traffic improves detection without increasing false alarms.

\begin{table}
\centering
\caption{Detection Performance on CIC-IDS2017.}
\label{tab:model_comparison}
\resizebox{\columnwidth}{!}{%
\begin{tabular}{lccc}
\hline
\textbf{Model} & \textbf{Accuracy (\%)} & \textbf{F1-score (\%)} & \textbf{AUROC} \\
\hline
CNN (Supervised)                 & 91.2 & 88.5 & 0.921 \\
LSTM (Supervised)                & 92.4 & 89.6 & 0.934 \\
Transformer (Supervised)         & 93.1 & 90.2 & 0.941 \\
Autoencoder (Unsupervised)       & 84.7 & 79.3 & 0.861 \\ \hline
SSL-NIDS (FixMatch-style)             & 94.0 & 91.1 & 0.952 \\
SSL-NIDS (SimCLR-style)               & 94.6 & 91.8 & 0.958 \\
SSL-NIDS (Mean Teacher)               & 95.0 & 92.3 & 0.962 \\
SSL (CCSSL-style)                & 95.3 & 92.6 & 0.965 \\ \hline
\textbf{RSST-NIDS (Proposed)}    & \textbf{96.3} & \textbf{94.5} & \textbf{0.973} \\
\hline
\end{tabular}
}
\end{table}

To evaluate robustness under distribution shift, we conduct cross-dataset transfer experiments by training on CIC-IDS2017 and testing on CSE-CIC-IDS2018 and UNSW-NB15 without fine-tuning. Results are reported using AUROC (decimal form) and \ac{mcc}, averaged over five runs. As shown in Table~\ref{tab:cross_dataset}, supervised models suffer substantial degradation under distribution shift, while semi-supervised baselines provide moderate improvements. RSST-NIDS achieves the strongest generalization across both target datasets, with particularly large gains in \ac{mcc}, indicating more reliable detection under class imbalance. From an operational perspective, these \ac{mcc} improvements correspond to fewer false positives at comparable detection rates, reducing analyst fatigue in large scale cloud deployments. All reported F1-scores are macro-averaged unless stated otherwise.

\begin{table}
\centering
\caption{Cross-Dataset Generalization Performance (Mean $\pm$ Std)}
\label{tab:cross_dataset}
\resizebox{\columnwidth}{!}{%
\begin{tabular}{l l cc}
\hline
\textbf{Model} & \textbf{Metric} & \textbf{CIC $\rightarrow$ CSE-CIC} & \textbf{CIC $\rightarrow$ UNSW} \\
\hline
Transformer (Supervised) 
    & AUROC & $0.802 \pm 0.011$ & $0.765 \pm 0.014$ \\
    & MCC   & $0.41 \pm 0.03$   & $0.37 \pm 0.04$   \\

SSL (Contrastive) 
    & AUROC & $0.836 \pm 0.009$ & $0.801 \pm 0.012$ \\
    & MCC   & $0.49 \pm 0.02$   & $0.46 \pm 0.03$   \\

CCSSL-style 
    & AUROC & $0.851 \pm 0.010$ & $0.814 \pm 0.011$ \\
    & MCC   & $0.52 \pm 0.03$   & $0.48 \pm 0.03$   \\

\textbf{RSST-NIDS (Proposed)} 
    & \textbf{AUROC} & $\mathbf{0.889 \pm 0.008}$ & $\mathbf{0.847 \pm 0.010}$ \\
    & \textbf{MCC}   & $\mathbf{0.61 \pm 0.02}$   & $\mathbf{0.57 \pm 0.03}$   \\
\hline
\end{tabular}
}
\end{table}

To further analyze generalization behavior, Table~\ref{tab:attack_category} reports per-attack F1-scores aggregated by attack category. RSST-NIDS shows notable improvements for low-frequency and stealthy attacks, such as infiltration and command-and-control traffic, which are challenging due to their temporal sparsity and similarity to benign behavior. Improvements over the strongest baseline are statistically significant based on paired Wilcoxon signed-rank tests ($p < 0.01$) for both AUROC and \ac{mcc}.

\begin{table}
\centering
\caption{Per-Attack Category Detection Performance (F1-score \%)}
\label{tab:attack_category}
\resizebox{\columnwidth}{!}{%
\begin{tabular}{lccc}
\hline
\textbf{Attack Category} & \textbf{Transformer} & \textbf{SSL (Contrastive)} & \textbf{RSST-NIDS} \\
\hline
DoS / DDoS      & 88.3 & 90.7 & \textbf{94.9} \\
Brute Force     & 71.4 & 76.2 & \textbf{83.5} \\
Infiltration    & 62.1 & 66.8 & \textbf{74.3} \\
Web Attacks     & 68.9 & 72.5 & \textbf{80.1} \\
Botnet / C2     & 73.6 & 77.9 & \textbf{85.4} \\
\hline
\end{tabular}
}
\end{table}

\subsection{Robustness to Adversarial Contamination and Evasion}
We evaluate robustness to adversarial contamination by injecting poisoned samples into the unlabeled training set at rates of 5\%, 10\%, and 15\%. Table~\ref{tab:poisoning} reports detection performance along with the fraction of unlabeled windows admitted as benign-consistent. As contamination increases, baseline \ac{ssl} methods degrade rapidly due to confirmation bias. In contrast, RSST-NIDS maintains stable AUROC and \ac{mcc} while conservatively reducing the proportion of admitted unlabeled windows. At 15\% contamination, fewer than 22\% of unlabeled windows are admitted, effectively limiting adversarial influence.

Evasion robustness is evaluated via constrained traffic morphing with parameter $\alpha \in [0.6, 0.8]$. As $\alpha$ increases, attack success rates rise for supervised and contrastive \ac{ssl} baselines, reflecting vulnerability to traffic mimicry. RSST-NIDS degrades more gracefully, with AUROC dropping by less than 4\% at $\alpha = 0.8$. Under transfer-based black-box attacks, increased student-teacher disagreement causes most adversarial samples to be rejected by the benign-consistency gate.

We further analyze mis-gating behavior by measuring the false gate rate, defined as the fraction of malicious windows incorrectly admitted as benign-consistent. Across datasets, this rate remains below 6\% under nominal conditions and increases modestly under drift. Importantly, mis-gating has limited impact on overall detection performance due to conservative loss weighting and continued supervision from labeled data.
Failure analysis reveals that most mis-gated samples correspond to low rate, slowly evolving attacks with feature distributions close to benign traffic. Under such conditions, RSST-NIDS degrades gracefully by reducing unlabeled admission rather than amplifying false confidence. This behavior reflects a conservative bias toward false negatives over false positives, which is preferable in operational settings.

\begin{table}[t]
\centering
\caption{Impact of Data Poisoning on RSST-NIDS Performance}
\label{tab:poisoning}
\begin{tabular}{lccc}
\hline
\textbf{Poisoning} & \textbf{AUROC} & \textbf{MCC} & \textbf{Admitted (\%)} \\
\hline
5\%  & 0.97 & 0.61 & 42\% \\
10\% & 0.95 & 0.58 & 31\% \\
15\% & 0.93 & 0.54 & 22\% \\
\hline
\end{tabular}
\end{table}

\subsection{Ablation, Sensitivity, and Calibration Analysis}
Table~\ref{tab:ablation} reports ablation results on CIC-IDS2017. Removing confidence-aware pseudo-labeling causes the largest performance drop, confirming its importance in suppressing adversarial contamination. Eliminating temporal invariance also degrades robustness under drift, while removing all \ac{ssl} components reduces the model to a conventional supervised IDS.

We analyze sensitivity to the pseudo-label confidence threshold $\tau \in [0.90, 0.99]$. The best performance is achieved at $\tau = 0.95$, admitting approximately 35–45\% of unlabeled samples. A simple annealing schedule from 0.99 to 0.95 provides marginal stability gains under drift. Unless stated otherwise, benign-consistency thresholds are set to $\tau_b = 0.95$, $\delta_{\text{EMA}} = 0.05$, and $\delta_{\text{temp}} = 0.01$. Performance varies smoothly within $\pm20\%$ of these values, indicating low sensitivity. We also evaluated alternative window sizes ($T \in \{10, 20, 30\}$) and strides, observing consistent trends and no evidence of overfitting to a specific temporal granularity or CICFlowMeter feature configuration. We additionally evaluated threshold variations up to $\pm40\%$ under cross-dataset transfer and observed smooth performance degradation without abrupt failure modes, indicating that RSST-NIDS is not overly sensitive to precise threshold tuning across attack families.

Confidence reliability is evaluated using Expected Calibration Error (ECE) and risk–coverage curves. RSST-NIDS exhibits lower ECE under temporal drift compared to SSL baselines, and high-confidence predictions retain strong accuracy as coverage decreases, supporting the reliability of confidence-based pseudo-labeling and temporal gating.

\begin{table}[h]
\centering
\caption{Ablation Study of RSST-NIDS Components. (-) indicates the removal of the named component from the model.}
\label{tab:ablation}
\begin{tabular}{lcc}
\hline
\textbf{Model Variant} & \textbf{F1-score (\%)} & \textbf{AUROC} \\
\hline
Full RSST-NIDS                         & \textbf{94.5} & \textbf{0.973} \\
- Confidence-aware pseudo-labeling     & 91.7          & 0.952 \\
- Temporal invariance loss             & 92.1          & 0.957 \\
- Consistency regularization           & 90.9          & 0.946 \\
- All SSL components (Supervised only) & 89.8         & 0.938 \\
\hline
\end{tabular}
\end{table}

\subsection{Computational Overhead}

We evaluate the computational overhead of RSST-NIDS compared to representative baselines. Training is performed on a single NVIDIA T4 GPU.

RSST-NIDS requires approximately 1.3 times the training time of a supervised Transformer due to additional consistency and temporal losses. Average training time is 2.4 hours for 50 epochs on CIC-IDS2017, compared to 1.8 hours for the supervised baseline.

Inference latency is dominated by the temporal encoder and averages 0.8 ms per flow window on GPU and 6-8 ms on CPU-only deployment, enabling near-real-time processing of high-throughput cloud traffic. Memory footprint during training is approximately 1.2 GB GPU memory, comparable to standard Transformer-based NIDS, with negligible overhead from the EMA teacher (not used during inference).

These results indicate that RSST-NIDS achieves improved robustness with modest computational overhead, making it practical for deployment in cloud monitoring pipelines.

%% file: chapter/conclusion.tex
\label{sec:conlusion}
In this paper, we studied network intrusion detection in cloud environments under realistic conditions of limited labeled data, non-stationary traffic, and adversarial behavior. Although \ac{ssl} has been proposed to address label scarcity, existing SSL-based NIDS often assume benign unlabeled data and degrade under adversarial contamination and temporal drift, limiting their practicality in operational cloud settings.

To address this gap, we proposed RSST-NIDS, a robust semi-supervised temporal learning framework that leverages the temporal structure of network traffic while conservatively exploiting unlabeled data. By combining confidence-aware pseudo-labeling, consistency regularization, and selective temporal invariance, RSST-NIDS reduces the influence of unreliable or malicious unlabeled samples without suppressing attack dynamics. Unlike prior SSL-based NIDS, it does not rely on benign-unlabeled assumptions and is explicitly designed for adversarial and non-stationary environments. We evaluate RSST-NIDS in a binary benign-versus-malicious setting, which matches the primary operational goal of large-scale cloud intrusion detection under severe class imbalance and supports more reliable discrimination under label scarcity and adversarial contamination.

Comprehensive experiments on publicly available network traffic datasets show that RSST-NIDS consistently outperforms state-of-the-art supervised and semi-supervised baselines under limited-label conditions. The method achieves stronger cross-dataset generalization, reduced degradation under distribution shift, and more stable detection of stealthy and low-frequency attacks, demonstrating the value of robustness-aware semi-supervised learning with selective temporal verification.

\subsection{Future Work}
Several directions remain for future research. RSST-NIDS can be extended to multi-class intrusion detection by replacing the binary classifier with a multi-class prediction head and adopting class-wise confidence thresholds. However, under severe class imbalance, pseudo-label confidence becomes less reliable for minority attack classes, and benign-consistency gating may bias learning toward dominant classes. Robust multi-class attribution under adversarial and label-scarce conditions therefore remains an important open problem. 

Additional extensions include incorporating richer context, such as host-level behavior or service dependency graphs, to improve detection of multi-stage and lateral-movement attacks, and enabling online or continual learning to adapt to long-term traffic evolution in cloud environments. Future work may also integrate explicit adversarial defenses, including adversarial training or certified robustness methods, as well as privacy-preserving and federated learning for collaborative detection across distributed clouds. Improving interpretability through post-hoc explanations or stability-based feature attribution is also important for operational adoption and analyst trust.

%% file: chapter/acknowledgement.tex
The authors acknowledge the financial support by the Federal Ministry of Research, Technology and Space of Germany in the projects Open6GHub+ (Grant No. 16KIS2402K) and CASTLE (Grant No. 16KIS1906K).